\documentclass[letterpaper, 10 pt, conference]{ieeeconf}  

\IEEEoverridecommandlockouts  
\usepackage{amsmath,amsfonts}
\usepackage{algorithm}
\usepackage{array}
\usepackage[caption=false,font=normalsize,labelfont=sf,textfont=sf]{subfig}
\usepackage{textcomp}
\usepackage{stfloats}
\usepackage{url}
\usepackage{verbatim}
\usepackage{graphicx}
\usepackage{cite}
\usepackage{xcolor}
\usepackage{adjustbox}
\usepackage{ragged2e}
\usepackage{tabularx}
\usepackage{multirow}
\usepackage{booktabs}
\usepackage{colortbl}
\usepackage{makecell}
\usepackage{arydshln}  
\usepackage[colorlinks,
linkcolor=blue,
anchorcolor=blue,
citecolor=blue]{hyperref}
\usepackage{marvosym}
\usepackage[capitalise]{cleveref}
\crefname{section}{Sec.}{Secs.}
\Crefname{section}{Section}{Sections}
\Crefname{table}{Table}{Tables}
\crefname{table}{Tab.}{Tabs.}

\usepackage{booktabs}
\usepackage[table]{xcolor}
\usepackage{amsmath, amssymb}
\usepackage{algpseudocode}
\usepackage{arydshln} 
\usepackage{multirow}
\usepackage{xcolor}

\newcommand{\oursname}{RoSe-SLAM}
\definecolor{lightgreen}{RGB}{235,242,202}
\definecolor{lightyellow}{RGB}{255,250,180}
\definecolor{softgreen}{RGB}{197,224,180}
\usepackage{bbm}
\usepackage{dsfont}
\usepackage{lmodern}
\usepackage{times}

\title{\oursname: Robust Semantic-Aware Gaussian Splatting SLAM from Dynamic Monocular Videos}

\author{Wenting Wang$^{1}$, Jiaxin Guo$^{1,\dagger}$, Wenzhen Dong$^{1}$, Yun-Hui Liu$^{1}$, Charlie C.L. Wang$^{2}$, Yeung Yam$^{1,3,\dagger}$
\thanks{$^{1}$ Department of Mechanical and Automation Engineering, The Chinese University of Hong Kong, Hong Kong SAR.}%
\thanks{$^{2}$ Department of Mechanical, Aerospace and Civil Engineering, The University of Manchester, Manchester, UK.}%
\thanks{$^{3}$ Centre for Perceptual and Interactive Intelligence (CPII) Limited, Hong Kong SAR.}%
\thanks{$\dagger$ Corresponding author:Yeung Yam(yyam@mae.cuhk.edu.hk), Jiaxin Guo (jiaxinguo001@cuhk.edu.hk)}
}

\begin{document}



\maketitle
\thispagestyle{empty}
\pagestyle{empty}





\begin{abstract}


In dynamic and unstructured environments, conventional SLAM systems generally suffer from significant accuracy degeneration due to their static assumptions. 
In this work, we propose \textit{Ro}bust \textit{Se}mantic-aware Gaussian Splatting \textit{SLAM} (\textit{RoSe-SLAM}), to address the dynamic challenge by a holistic semantic scene understanding from \textit{uncalibrated monocular inputs}, achieving accurate camera tracking and high-quality geometry reconstruction.
Unlike conventional semantic SLAM using handcrafted semantic labels, our \oursname{} exploits the semantic feature from 2D foundation model to enhance the dynamic tracking and mapping performance. 
By distilling the rich semantic features to our Gaussian fields, our method effectively identifies dynamic distractors and achieves semantic-aware multi-view consistency, significantly enhancing the geometric reconstruction and scene inpainting. 
Specifically, we propose a \textbf{spatial-temporal motion mask generation module}, enabling both long-term motion monitoring and short-term transient dynamics capturing, achieving robust and effective disentanglement of dynamic objects and static backgrounds.
%
During global bundle adjustment, we propose an \textbf{occlusion-aware keyframe selection mechanism} to prioritize the occlusion as metric to pick the keyframes, and a \textbf{multi-view semantic consistency module} to improve the mapping quality in dynamic environments. By combining geometric motion cues with semantic priors, our system dynamically filters unreliable observations and reconstructs accurate static scene geometry. 
Extensive experiments conducted on benchmark datasets including dynamic TUM, Bonn and Wild-Mocap datasets, demonstrate that our method achieves superior performance in both trajectory estimation and static scene mapping, outperforming existing dynamic RGB SLAM baselines in long-term dynamic indoor environments. 

\end{abstract}
\section{Introduction}

Visual Simultaneous Localization and Mapping (SLAM) has been extensively investigated over decades and applied to broad real-world applications such as robotics, autonomous driving, and virtual/augmented reality~\cite{rosinol2020kimera, mccormac2017semanticfusion, lu2025ring, shi2024lidar_slam, luo2025bevplace++, guo2025salon3r}. 
However, most of the approaches~\cite{campos2021orb, teed2021droidslam} are highly dependent on the scene rigidity assumption, leading to significant performance degradation in dynamic scenes where objects move independently. Such conditions frequently cause tracking failures and chaotic mapping, limiting the applications in real-world scenes. Consequently, advancing SLAM techniques for dynamic scenes remains a critical research challenge. 


Despite recent progress in neural SLAM, most systems still degrade significantly in dynamic scenes with frequent distractors such as pedestrians, moving objects, and occlusions~\cite{sabour2023robustnerf, ren2024nerf-on-the-go}. 
This is largely due to the inherently dynamic nature of real-world environments, which renders the common assumption of static or distractor-free scenes impractical. Moreover, removing dynamic objects from captured data often requires dense per-pixel annotations, which are labor-intensive and difficult to scale to long sequences or large scenes. This reliance on costly supervision remains a key bottleneck for the practical deployment of SLAM in real-world applications.
Recently, several 3DGS-based SLAM systems have been proposed to address this challenge~\cite{xu2025dgslam, kong2024dgs-slam}, but they often rely heavily on depth input. In contrast, our method operates solely on monocular RGB video, without requiring camera intrinsics, poses, or depth, while achieving comparable accuracy and robustness. 
Recent work like WildGS-SLAM~\cite{zheng2025wildgs-slam} introduces monocular dynamic SLAM using uncertainty modeling, but struggles with relative depth from DBA~\cite{teed2021droidslam}, contributing to poor reconstruction quality. We instead jointly optimize a semantic Gaussian field to improve both scene robustness and completeness.

\begin{figure}[t]
    \centering
    \includegraphics[width=1.0\columnwidth]{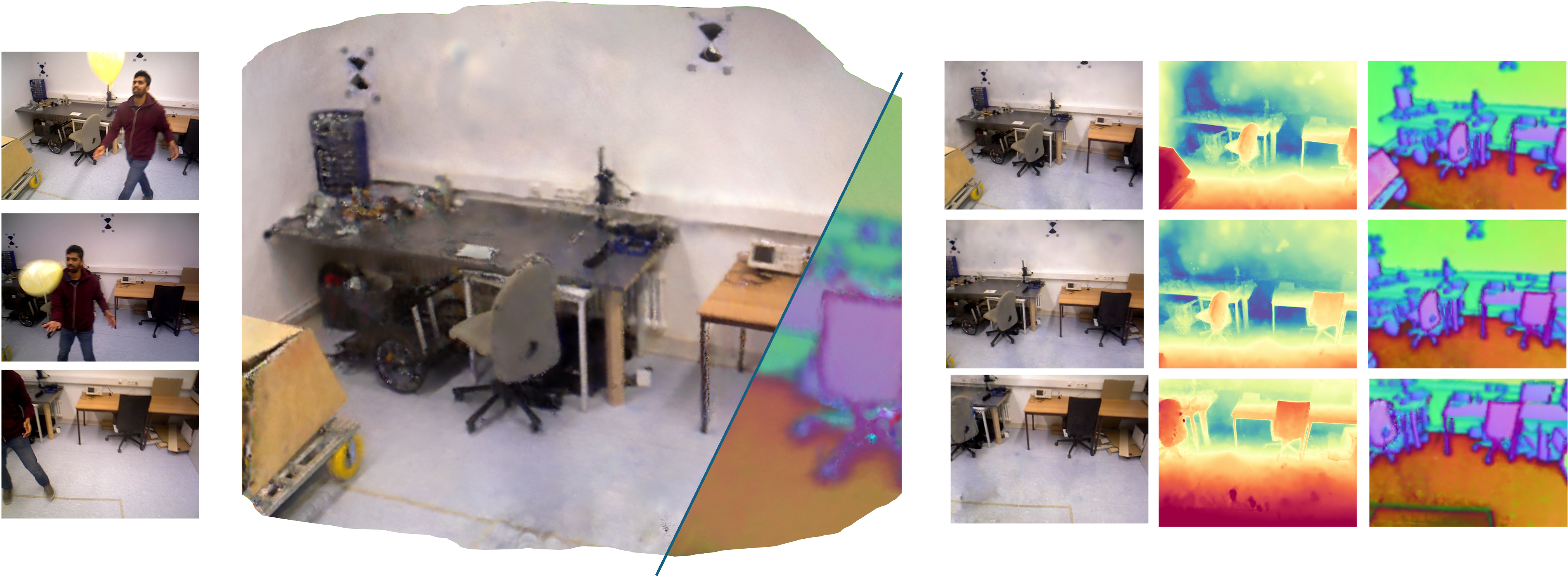} 
    \put(-245,-5){\scriptsize \color{black} Input video}
    \put(-200,-5){\scriptsize \color{black} Reconstructed static scene}
    \put(-95,-5){\scriptsize \color{black} Rendered RGB/Depth/Semantics}
    \caption{{\textbf{Our \textbf{RoSe-SLAM} addresses the dynamic reconstruction challenge via holistic semantic scene understanding from uncalibrated monocular inputs}, effectively removing distractors and mitigating geometry inconsistencies, achieving superior reconstruction performance.} 
    }
    \label{fig:teaser}
\end{figure}

To enhance the accuracy of mapping process in dynamic scenarios, we draw on human-like semantic intuition. Specifically, objects observed from different viewpoints are intuitively associated with consistent semantic features, which facilitates multi-view semantic consistency. Motivated by this insight, we propose learning a semantic Gaussian field tailored for dynamic environments by leveraging semantic features extracted from powerful foundation models such as \cite{ravi2024sam2, oquab2023dinov2}. This strategy aligns well with recent advancements in semantic 3D Gaussian Splatting methods \cite{qin2024langsplat, zhou2024feature-3dgs, zuo2024fmgs}, which enable tasks such as text-prompt-based segmentation, scene editing and manipulation, and geometry refinement \cite{qiu2024gls}. Within the context of dynamic SLAM, semantic fields can be smoothly incorporated into motion-consistent dynamic mask generation, enforcing multi-view semantic consistency, and guiding occlusion inpainting. Furthermore, the integration of semantic information not only benefits robust SLAM performance in dynamic scenes but also supports a range of downstream robotic applications, including perception, autonomous navigation, and scene-level understanding, thereby broadening the practical utility of the system in real-world robotic applications.

In this paper, we present \textbf{Ro}bust \textbf{Se}mantic-aware Gaussian Splatting \textbf{SLAM} (\textbf{\oursname{}}), to address the dynamic environment challenges using only \textit{uncalibrated monocular RGB input}, achieving accurate camera tracking and high-quality geometry reconstruction. 
Instead of using handcrafted semantic labels, we propose a \textbf{spatial-temporal motion mask generation module} to enable both long-term motion tracking and short-term dynamics detection, achieving robust and effective disentanglement of dynamic objects and static backgrounds.
During global bundle adjustment, we propose an \textbf{occlusion-aware keyframe selection} mechanism that considers occlusion as a prior metric to pick keyframes, improving both rendering quality and scene completeness. To mitigate the geometric inconsistency across multiple views, we propose a \textbf{multi-view semantic consistency module} to enhance the accuracy and robustness of camera pose estimation and mapping. This semantic-aware multi-view consistency strengthens geometric coherence in the absence of reliable depth, while also mitigating occlusions and missing geometry through semantic-guided occlusion supervision. 
%

Our \textbf{contributions} are summarized as follows: 
\begin{enumerate}
    \item We present a dynamic dense SLAM system that leverages semantic features for enhanced geometric consistency from \textit{uncalibrated monocular RGB input}.
    \item A \textbf{spatial-temporal motion mask generation} module is introduced to enable both long-term motion tracking and short-term dynamics detection, achieving robust and effective disentanglement of dynamic objects and static backgrounds.
    \item During global bundle adjustment, an \textbf{occlusion-aware keyframe selection} mechanism is introduced to facilitate occlusion region inpainting. We further propose a \textbf{multi-view semantic consistency} module to enhance the accuracy and robustness of camera pose estimation and mapping. 
    \item Extensive evalutions on challenging real-world dynamic datasets demonstrate the state-of-the-art performance of our proposed dynamic RGB SLAM. 
\end{enumerate}
\section{Related Work}

\noindent\textbf{Visual SLAM.}
Visual dynamic SLAM methods fall into two categories. The first relies on re-sampling and residual optimization to remove outliers, as in ORB-SLAM2 \cite{mur2017orbslam2}, ORB-SLAM3 \cite{campos2021orbslam3}, and Refusion \cite{palazzolo2019refusion}. However, these approaches are limited to small-scale motions and struggle with sustained, large dynamic changes.
The second category leverages priors like segmentation or object detection to filter dynamic content. Though more adaptable, they often suffer from domain gaps and prediction errors in real-world deployment.
End-to-end learning-based approaches such as DROID-SLAM \cite{teed2021droidslam}, DytanVO \cite{shen2023dytanvo}, and DeFlowSLAM \cite{ye2022deflowslam} improve motion estimation in dynamic scenes but require large-scale training data and struggle to produce high-fidelity static maps.
We address the long-term dynamic problem by modeling the scene with 3D Gaussian Splatting fused semantics, which can robustly represent both static and dynamic elements.

\noindent\textbf{3DGS SLAM.} 
NeRF-based SLAM methods \cite{mildenhall2021nerf} have shown strong scene representation with low memory usage, e.g., iMAP \cite{sucar2021imap}, NICE-SLAM \cite{zhu2022niceslam}, and Point-SLAM \cite{sandstrom2023pointslam}. However, they assume static scenes and degrade with dynamic content.
Rodyn-SLAM \cite{jiang2024rodynslam} attempts to address dynamics using optical flow and semantic priors, but its neural rendering frontend is computationally heavy and limits pose accuracy.
In contrast, 3D Gaussian Splatting (3D-GS) \cite{kerbl2023_3dgs} offers real-time, high-fidelity rendering by explicitly modeling scenes as Gaussian ellipsoids. Some recent efforts explore replacing implicit NeRFs with 3DGS in SLAM pipelines, though most still assume static environments.
DG-SLAM \cite{xu2025dgslam} introduces 3DGS into dynamic SLAM but depends on DROID-SLAM for pose and depth initialization. 
Our method accepts monocular input and deal with long-term dynamics, enhancing robustness and practicality.

\noindent\textbf{Semantic 3DGS.}
Semantic understanding is essential for SLAM applications in robotics and AR/VR. Earlier systems like CodeSLAM \cite{bloesch2018codeslam}, SLAM++ \cite{salas2013slam++}, and Kimera \cite{rosinol2020kimera} integrate semantics into sparse 3D geometry using voxels \cite{hermans2014dense_semantic}, point clouds \cite{narita2019panopticfusion}, or signed distance fields. These methods often lack reconstruction fidelity, speed, and memory efficiency.
SGS-SLAM \cite{li2024sgs-slam} introduces SAM-derived semantic features into 3D Gaussians, but is still constrained to static scenes. In contrast, our method targets dynamic environments, enabling robust semantic SLAM in the presence of long-term moving people and objects.

\section{Method}
\begin{figure*}[t]
    \centering
    \includegraphics[width=1.0\textwidth]{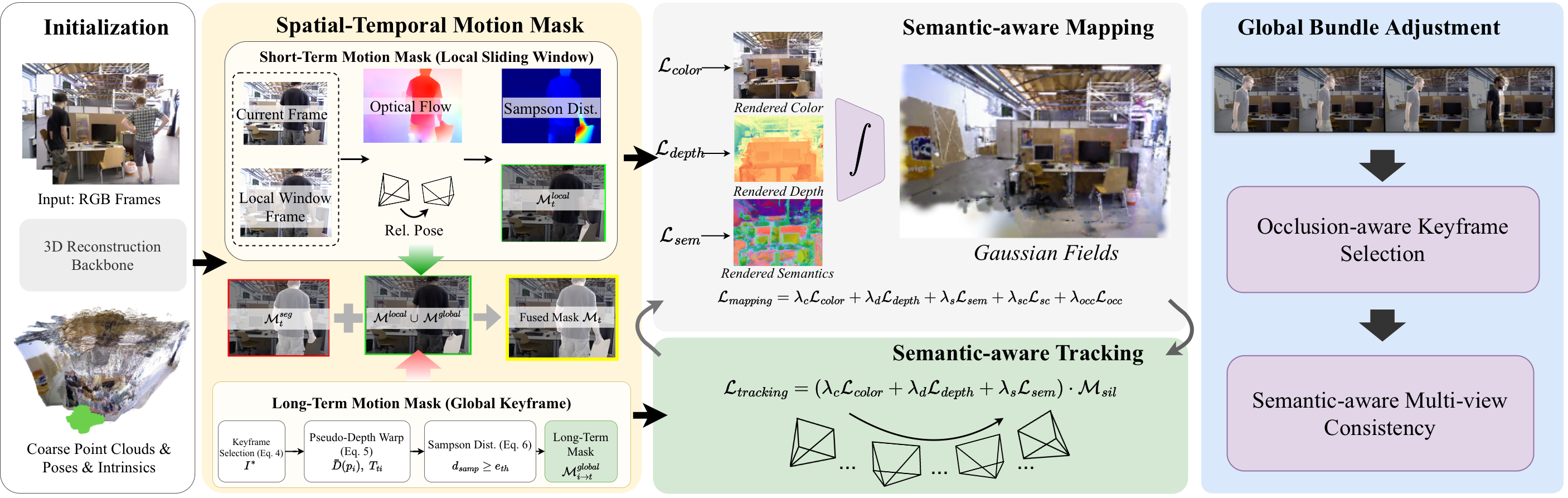}  
    \caption{\textbf{Overview of RoSe-SLAM.} Given a sequence of RGB frames, we propose a robust semantic-aware Gaussian Splatting SLAM to jointly estimate the semantic representation and camera poses, from \textit{uncalibrated monocular inputs}. To address the dynamic challenge, a spatial-temporal motion mask generation module is employed to achieve effective disentanglement of dynamic objects and static backgrounds. During global bundle adjustment, an occlusion-aware keyframe selection module and multi-view semantic consistency module are introduced to improve the inpainting and geometry quality. }
    \label{fig:method}
\end{figure*}
We propose RoSe-SLAM, a robust long-term dynamic visual SLAM system with semantic 3D Gaussians, aiming to achieve robust pose estimation and reconstruction under the challenging dynamic scene.

\subsection{Semantic 3D Gaussian Representation}
Our goal is to establish a scene representation that effectively captures the high-fidelity appearance, geometry, and semantics of the static region within the dynamic real-world scenes. To this end,we filter out dynamic elements and represent the static scene using a set of semantic 3D Gaussians: 
\begin{equation}
    \mathcal{G} := \{ (\mu_i, \Sigma_i, o_i, c_i, f_i) \mid i = 1,2,\dots,N \}.
    \label{eq:gs}
\end{equation}
Upon the projection of 3D Gaussians onto the image plane, the color $\hat{c}$ and depth $\hat{d}$ is rendered by sorting the Gaussians in depth order and performing 
front-to-back \( \alpha \)-blending rendering as:

\begin{equation}
    \hat{c} = \sum_{i}^{M} c_i \alpha_i \prod_{j}^{i-1} (1 - \alpha_j), \quad \hat{d} = \sum_{i}^{M} d_i \alpha_i \prod_{j}^{i-1} (1 - \alpha_j),
    \label{eq:gs_color_pix}
\end{equation}
where \( M \) is the number of sorted Gaussians overlapping with the given pixel. The density 
\( \alpha_i \) is computed from the 2D covariance matrix \( \Sigma_{2D, i} \) and the opacity 
\( o_i \) of the \( i \)-th 3D Gaussian by $\alpha_i = o_i \cdot \exp \left( -\frac{1}{2} \sigma_i^T ({\Sigma^{2D}_{i}})^{-1} \sigma_i \right)$, 
where \( \sigma_i \in \mathbb{R}^2 \) is the offset between the pixel center and the \( i \)-th 
projected 2D Gaussian center. 






We propose multimodal properties of Gaussian primitives, especially with semantics, to improve the multi-view geometry of the incremental reconstruction in dynamic environments. 
Similar to color and depth rendering, the semantic feature of a single pixel \( \hat{f}^{d_{\text{sem}}} \) can be rendered as, 
\begin{equation}
    \hat{f}^{d_{\text{sem}}} = \sum_{i}^{M} f_i^{d_{\text{sem}}} \alpha_i \prod_{j}^{i-1} (1 - \alpha_j).
    \label{eq:gs_semantic_pix}
\end{equation}
However, SLAM systems generally process long sequence frames while semantic features typically have high-dimensional channels. Directly rendering high-dimensional features is impractical due to excessive time consumption and storage cost. 
To catch up with high rendering speed and comparative space cost, we render a low-dimensional feature \( d_{\text{sem}} \) first and a decoder is employed to obtain the high-dimensional \( D_{\text{sem}} \) semantic feature of a single pixel \( \hat{f}^{D_{\text{sem}}} =  \text{Decoder} \left( \hat{f}^{d_{\text{sem}}} \right) \). 
Our approach possesses the capability to render semantic feature maps of varying dimensions by using an auto-encoder. 
Specifically, we employ \cite{ravi2024sam2} semantic features as supervision and follow \cite{qin2024langsplat} to train an auto-encoder and decoder framework to obtain embedded semantic features with low dimensionality as the initialization of 3D Gaussians, and pass the rendered low-dimensional features through the pre-trained decoder to extend to high-dimensional features. 


\subsection{Coarse Initialization}
Following \cite{wang2025cut3r}, given RGB input, we obtain coarse depths, camera poses, and intrinsics for each image. However, these depths and camera parameters serve as a coarse initialization, which is not as precise as the ground truth and consists of noise. 
To enhance the monocular depth with both fine-grained details and consistent scale, we prompt a metric depth foundation model~\cite{lin2025promptda} with scale-consistent depths and then obtain the fine and detailed depth as each frame's pseudo-depth $\tilde{D}$. 


\subsection{Spatial-Temporal Motion Mask}
To filter out the movable objects in dynamic environments, current methods either utilize optical flow accumulated on several consecutive frames and Sampson distance threshold \cite{jiang2024rodynslam} or employ depth-warping residuals \cite{xu2025dgslam} within a sliding window to generate motion masks. However, there are objects that cannot be detected by these approaches because they are static in consecutive frames but movable for a long time. For example, there is a chair that someone is moving in the end but static in the middle of the input videos, so that in the middle frames these approaches cannot mask the objects.
We propose a spatial-temporal adaptive motion mask generation to produce movable object masks from long-term and overlapped frames in a global perspective.

 Warping methods based on local sliding windows focus on the motion within a short time, while they neglect the long-term motion. Specifically, we design a strategy to select the keyframe candidates to be aligned with the current frame to generate the motion masks not only in a local sliding window but also for a long time. We select the keyframe that has enough overlap and meanwhile keeps as long-time as possible, following the equation as:
\begin{equation}
    I^* = \arg\max_{I_i} \lambda \cdot O(I_t, I_i) + (1 - \lambda) \cdot \Delta T(i, t),
\end{equation}
where $O(\cdot, \cdot)$ represents the spatial overlap of two frames, $\Delta T(\cdot, \cdot)$ denotes the time interval between keyframe $i$ and current frame $t$. $\lambda$ is the balance parameter defining which condition has a higher weight to select a keyframe. $I_t$ is the current frame while $I_i$ is one of keyframes belong to the keyframe set. 

\begin{equation}
p_{i \rightarrow j}
= \pi\!\left( K_j \left( R_{ji}\, \tilde{D}(p_i)\, K_i^{-1}\, p_i^{\mathrm{homo}} + t_{ji} \right) \right),
\end{equation}
where $K_i$ and $K_j$ denote the camera intrinsics of frame $i$ and frame $j$,
$T_{ji} = [R_{ji} \mid t_{ji}] \in \mathrm{SE}(3)$ is the relative pose from frame $i$ to frame $j$,
and $\pi([x, y, z]^{\top}) = [x/z,\, y/z]^{\top}$ denotes the perspective projection.
$\tilde{D}(p_i)$ denotes the pseudo depth of pixel $p_i$. 
\( {p}_i^{\text{homo}} = (u, v, 1) \) is the homogeneous coordinate.

We utilize the optical flow and Sampson distance error to produce the motion. To separate the ego-motion from dynamic objects, we additionally estimate the fundamental matrix \( {F} \) with inliers sampled from the matching set \( S \). Given any matching points \( {o}_{i \to t} \) within \( S \), we utilize matrix \( {F} \) to compute the Sampson distance between corresponding points and their epipolar lines. By setting a suitable threshold \( e_{\text{th}} \), we derive the warp mask corresponding to dynamic objects as:
\begin{equation}
M_{i \rightarrow t}^{\mathrm{global}}(p) \;=\;
\mathbbm{1}\!\left[\, d_{\mathrm{samp}}\!\left(F,\; p,\; o_{i \rightarrow t}(p)\right) \,\geq\, e_{\mathrm{th}} \,\right],
\label{eq:global_mask}
\end{equation}
where $\mathbbm{1}[\cdot]$ is the indicator function,
$d_{\mathrm{samp}}(F, p, p')$ measures the Sampson distance of the
correspondence $(p, p')$ 
and $e_{\mathrm{th}}$ is the inlier threshold.
$M_{i \rightarrow t}^{\mathrm{global}}(p) = 1$ indicates that $p$ violates
the epipolar constraint of the static background and is regarded as dynamic.
The final motion mask $M_t$ is defined as:
\begin{equation}
    M_{t}=M_{i \to t}^{\text{global}} \cup M_{\mathcal{J} \to t}^{\text{local}} \cup M_{t}^{\text{seg}},
\end{equation}
where $i$ denotes the keyframe selected by Eq.~(4),
$\mathcal{J}$ is the set of frames within the local sliding window,
$M_{\mathcal{J} \rightarrow t}^{\mathrm{local}}$ is computed following Eq.~(6)
for each frame $j \in \mathcal{J}$,
and $M_{t}^{\mathrm{seg}}$ is the segmentation-based motion mask.
\subsection{Bundle Adjustment}
\subsubsection{Occlusion-aware Keyframe Selection}
In conventional visual SLAM systems utilizing a sliding window keyframe buffer, a single keyframe is typically selected at random for mapping supervision. However, in dynamic environments with distractors, this strategy can lead to insufficient supervision of occluded regions, as such areas may not be consistently observed across randomly chosen keyframes, resulting in suboptimal optimization of regions that are frequently occluded or only partially visible. To solve this, increasing the probability of selecting keyframes where the occluded regions are clearly observed allows the system to better supervise and refine these areas from the current viewpoint. We propose a strategy to increase the possibility of the occluded areas being chosen. Specifically, we warp the masked areas in the current frame to each frame in keyframe buffer by using accordingly estimated camera parameters and depths, and the bigger area proportion of occlusion in each frame represents the larger possibility to be selected. We adopt a novel occlusion buffer to save the observed regions of occluded images. If the key frames in which the occluded region is visible are more likely to be selected, the occluded area can be more effectively optimized from the current viewpoint, thereby improving the overall inpainting performance.




\begin{equation}
\mathcal{L}_{\mathrm{occ}} \;=\;
\left\|
\left( \hat{I}_{t \rightarrow k} - I_k \right)
\odot \left( 1 - M_k \right)
\odot M_{t \rightarrow k}
\right\|_1,
\label{eq:L_occ}
\end{equation}

where $k$ denotes the selected keyframe,
$\operatorname{I}_{t \rightarrow k}(\cdot)$ warps an image from frame $t$
to the viewpoint of keyframe $k$ by applying Eq.~(5),
$M_k$ is the motion mask defined in Eq.~(7),
$M_{t \rightarrow k}$ is the
current motion mask warped into keyframe $k$,
$\odot$ denotes the element-wise product.

\subsubsection{Semantic-aware Multiple-view Consistency}

As advanced semantic foundation models \cite{oquab2023dinov2, ravi2024sam2, cherti2023openclip} have developed, recent approaches \cite{zhou2024feature-3dgs} incorporate semantic features with 3DGS to enrich properties of scenes, which is beneficial for surface reconstruction \cite{qiu2024gls} and downstream tasks such as editing and manipulation. However, occlusions and perspective changes introduce inconsistencies, causing the same pixel to be assigned different semantic interpretations across various viewpoints. To address these challenges, we utilize pixel embeddings that are both object-distinguishable and view-consistent. 



Using view-consistent semantics, we can enhance mapping performance since adjacent pixels within the same semantic category generally show small spatial distance variations. Intuitively, this property enables the identification of noise points and artifact floaters in 3D Gaussians. Another inspiration is that the same object in various images should maintain the same semantic features. Therefore, we warp the rendered semantic features $\hat{f_t}$ in image $I_t$ into previous frames $I_{t-1}$ by using the estimated camera transformation $T$,
and the distance between warped semantics $\hat{f}_{t \rightarrow t-1}$ and the rendered semantics $\hat{f}_{t-1}$ should be as small as possible in feature space,


\begin{equation}
\small
\mathcal{L}_{\mathrm{sc}}=
\frac{1}{\mathcal{T}} \sum_{t}^{\mathcal{T}}
\left\|
\left( \hat{f}_{t \rightarrow t-1} - \hat{f}_{t-1} \right)
\odot \left( 1 - M_{t \rightarrow t-1} \right)
\odot M_{t-1}^{\mathrm{sil}}
\right\|_1,
\label{eq:feature_wapr_loss}
\end{equation}
where $M_t^{\mathrm{sil}} = \mathbbm{1}\!\left[\textstyle\sum_{n=1}^{M}
\alpha_n \prod_{m=1}^{n-1}(1-\alpha_m) > \tau_{\mathrm{sil}}\right]$
is the silhouette mask indicating pixels with sufficient Gaussian coverage.



\subsection{Semantic-aware SLAM System}
Transient objects such as cars outdoors and books and chairs indoors commonly appear in daily life, while traditional multi-view geometry methods rely on static scenes to recover 3D structures. Therefore, conventional SLAM faces challenges in dynamic environments with distractors and humans. We propose employing high-level semantic features to enhance multi-view geometry consistency and improve the occluded background in-painting. The tracking and mapping processes are optimized in an interleaved manner.

\subsubsection{Mapping Process}
During mapping, the scene map is initialized from the first frame, and progressively adds new Gaussians in the newly observed areas and low-silhouette regions in the next frame, which accelerates the optimization process and meanwhile reduces the number of Gaussians, leading to time and storage efficiency. 
The scene is modeled using 3D Gaussians across three distinct channels: (1) 3D Gaussians coordinates depict the scene's geometry; (2) sphereical coefficients represent the visual appearance of the scene; (3) semantic features indicate the implicit category features and benefit robotic perception. These parameters are jointly optimized during the mapping process and while in the tracking process Gaussian parameters remain fixed.
Overall mapping loss is summarized as,

{\footnotesize
\begin{equation}
    \mathcal{L}_{color} = \frac{1}{ \mathcal{T}} \sum_{t}^{ \mathcal{T}} \left\| \left( \hat{C}_t - C_t \right) \odot \left(1-M_t \right) \right\|_2 + \lambda_{\mathrm{ssim}}
    \left( 1 - \mathrm{SSIM} \right),
    \label{eq:loss_rgb}
\end{equation}
\begin{equation}
    \mathcal{L}_{sem} = \frac{1}{\mathcal{T}} \sum_{t}^{\mathcal{T}} \left\| \left( \hat{f}_t - f_t \right) \odot \left(1-M_t \right) \right\|_1,
    \label{eq:loss_sem}
\end{equation}
\begin{equation}
     \mathcal{L}_{depth} = \frac{1}{\mathcal{T}} \sum_{t}^{\mathcal{T}} \left\| \left( \hat{D}_t - \tilde{D}_t \right) \odot \left( 1 - M_t \right) \odot \mathcal{W}_t \right\|_2,
\end{equation}
\begin{equation}
\begin{aligned}
     \mathcal{L}_{mapping} &= 
        \lambda_c \mathcal{L}_{color} + 
        \lambda_d \mathcal{L}_{depth} + 
        \lambda_s \mathcal{L}_{sem} \\ &+ 
        \lambda_{sc} \mathcal{L}_{sc} + 
        \lambda_{occ} \mathcal{L}_{occ}. 
\end{aligned}
    \label{eq:loss_mapping}
\end{equation}
}
where $\hat{D}_t$ is the rendered depth map,
$\tilde{D}_t$ is the pseudo-depth,
and $\mathcal{W}_t$ is the per-pixel confidence weight of the pseudo-depth.

\subsubsection{Tracking Process}
During tracking, we aim to fix the Gaussian parameters and optimize camera poses only. Following coarse-to-fine relative pose estimation, the current frame pose is iteratively refined by minimizing the tracking loss between the ground truth color, depth images and semantic maps and their differentially rendered views.
\begin{equation}
    \mathcal{L}_{tracking} = (
        \lambda_c \mathcal{L}_{color} + 
        \lambda_d \mathcal{L}_{depth} + 
        \lambda_s \mathcal{L}_{sem}
    ) 
    \cdot \mathcal{M}_{sil}.
    \label{eq:loss_tracking}
\end{equation}






\section{Experiments}

\noindent\textbf{Datasets.}
Our methodology is evaluated using publicly available challenging datasets: the TUM RGB-D dataset~\cite{sturm2012ate}, the Bonn RGB-D Dynamic dataset~\cite{palazzolo2019refusion} and Wild-Mocap dataset~\cite{zheng2025wildgs-slam}. These datasets encompass highly dynamic environments, including movable humans and objects, enabling a comprehensive evaluation of our approach under diverse conditions. This selection highlights the effectiveness and robustness of our method in real-world indoor scenarios.

\noindent\textbf{Metrics.} 
For the evaluation of pose estimation, we employ the Root Mean Square Error (RMSE) of the Absolute Trajectory Error (ATE) \cite{sturm2012ate}. To ensure a consistent basis for comparison, the estimated trajectory is first aligned with the ground truth trajectory using Horn’s Procrustes method \cite{horn1987hp}. Additionally, to assess the reconstruction quality of static maps in dynamic environments, we adopt three key metrics: (i) Accuracy (cm), (ii) Completion (cm), and (iii) Completion Ratio (the percentage of points within a 5 cm threshold), following the evaluation protocol of \cite{zhu2022niceslam}. We first cull the ground truth point cloud using training camera frastums, and then randomly sample 200,000 points using farthest sampling strategy from ground truth point cloud and the reconstructed mesh surface since the Bonn dataset offers only the ground truth point cloud. 


\noindent\textbf{Implementation details.}
We evaluate the performance of our method on an RTX 4090 GPU. 
We leverage \cite{ravi2024sam2} for prior semantic segmentation. For optical flow warp masking, we define a window size of 4 and a depth threshold of 0.6. Additionally, keyframes are selected following~\cite{teed2021droidslam}. We also employ \cite{wang2025cut3r} to produce coarse point clouds and camera parameters and utilize \cite{lin2025promptda} to produce finer depth priors.

\subsection{Evaluation of mapping performance}

\begin{figure}[htbp]
    \centering
    \includegraphics[width=1.0\linewidth]{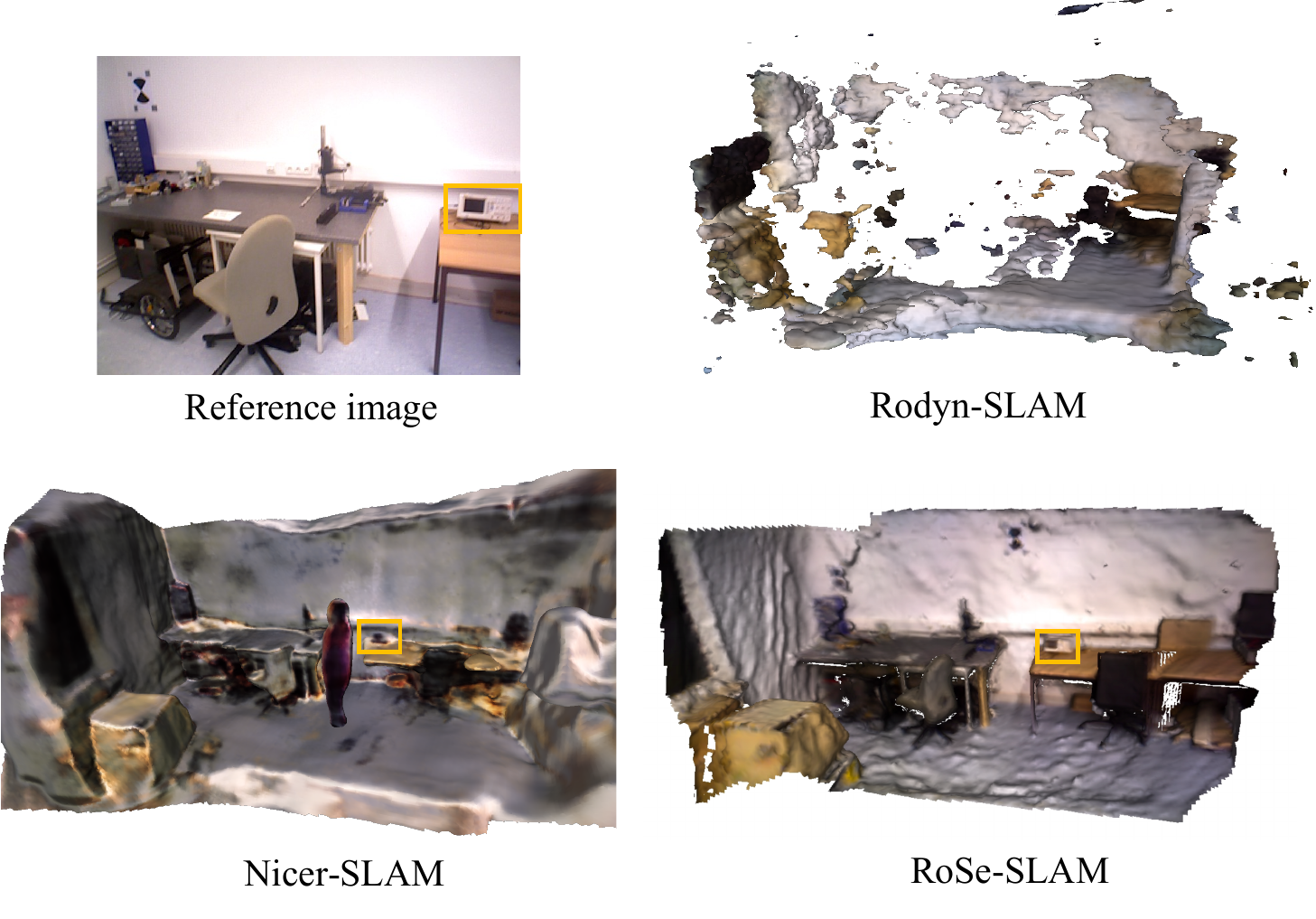}
    
    \caption{\textbf{Qualitative comparison for static scene reconstruction performance on Bonn dataset.} The proposed method achieves the best performance in reconstruction.}
    \label{fig:exp_recon}
\end{figure}

To more comprehensively assess the mapping performance of our method in dynamic scenes, we conduct both qualitative and quantitative evaluations of the reconstruction results. As dynamic scene datasets rarely offer static ground-truth meshes or point clouds, we adopt the Bonn dataset for our quantitative analysis. Our method is compared against state-of-the-art GS-based RGB and RGBD SLAM methods. As reported in Table~\ref{tab:recon_bonn}, our method significantly outperforms contemporary approaches across multiple metrics, including accuracy, completeness and completion ratio, achieving state-of-the-art performance. Furthermore, the reconstructed static maps exhibit high visual fidelity, as illustrated in Figure~\ref{fig:exp_recon}, further indicating that our method generates more accurate static reconstructions than prevailing SLAM systems.


\begin{figure}[t]
\vspace{1em}
    \centering
    \includegraphics[width=1.0\linewidth]{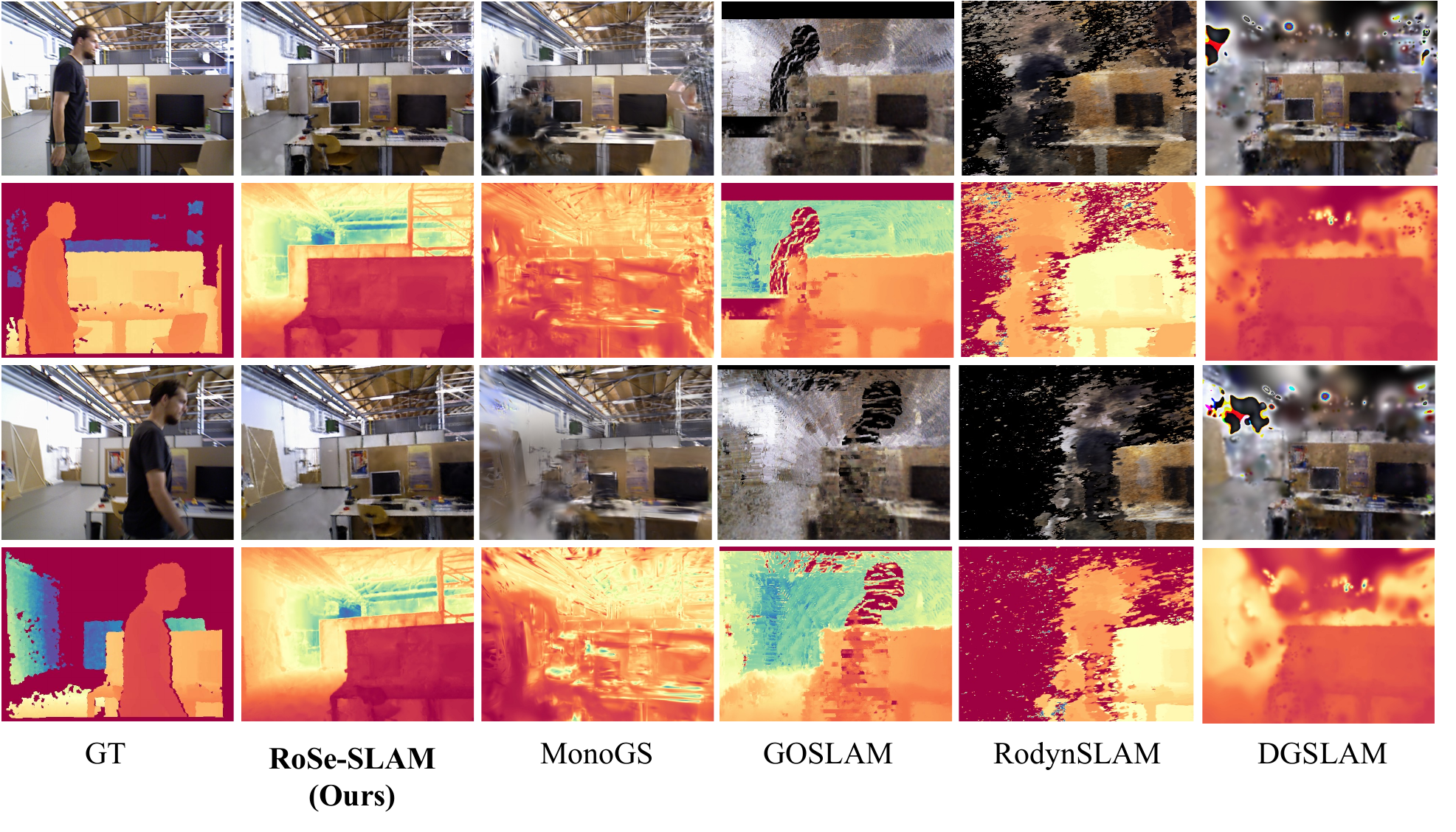}
    \caption{{\textbf{Qualitative comparison of rendered images and depths on TUM dataset}~\cite{sturm2012ate}.} Both the geometry and photometric results demonstrate the effectiveness of our method to deal with dynamic objects.}
    \label{fig:supp_tum}
\end{figure}

\begin{figure}[t]
    \centering
    \includegraphics[width=1.0\linewidth]{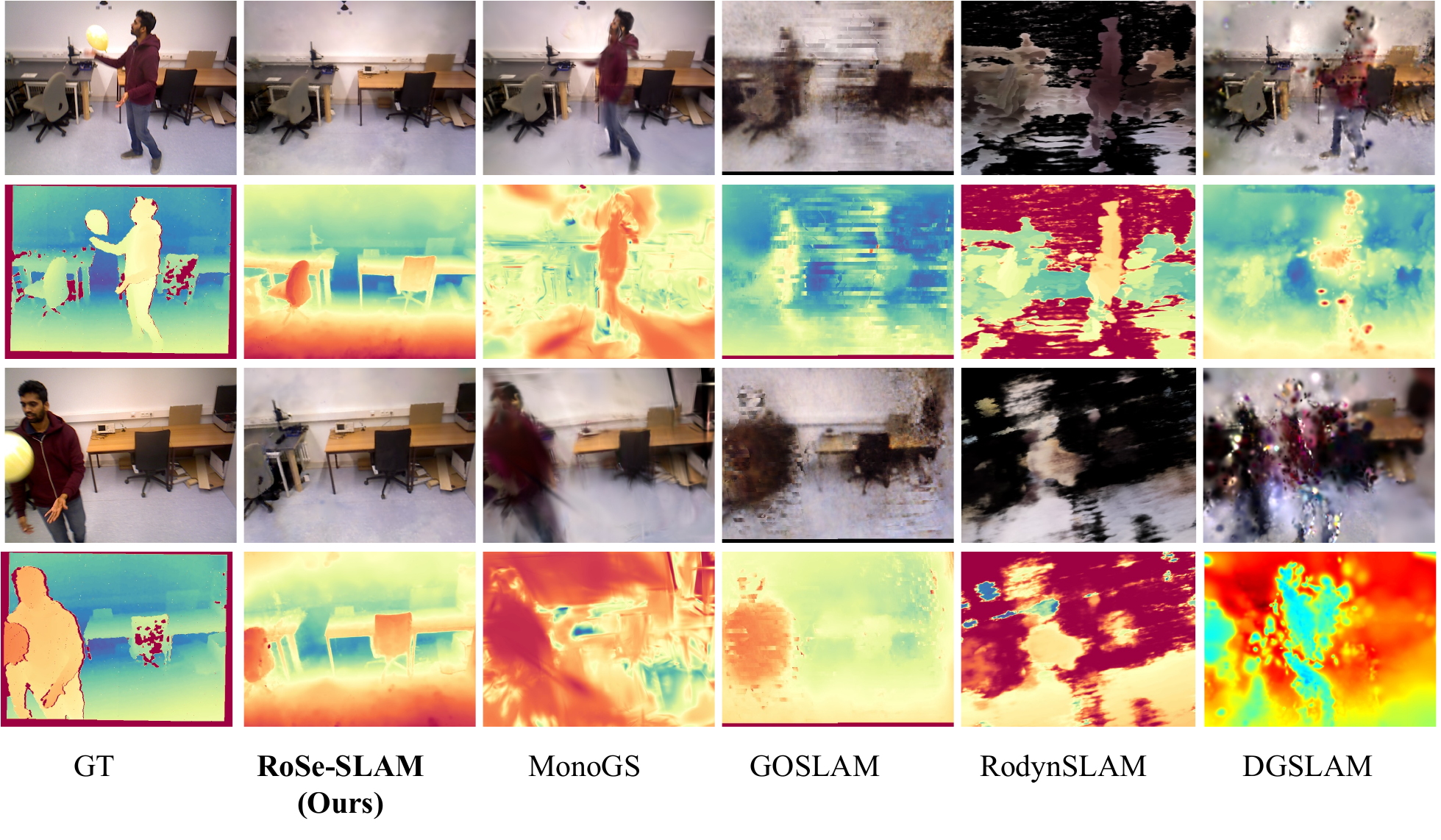}
    \vspace{-2em}
    \caption{{\textbf{Qualitative comparison of rendered images and depths on Bonn dataset~\cite{palazzolo2019refusion}.} Our Rose-SLAM is capable of removing distractors and occlusions, rendering high quality images without blurry artifacts.}}
    \label{fig:supp_bonn}
\end{figure}

\makeatletter
\def\heavyrulewidth{1.5pt}
\makeatother

\begin{table}[h]
    \centering
    \caption{Reconstruction results on several dynamic scene sequences in the  {BONN} dataset. Instances of tracking failures are denoted by ``X''.}
    \label{tab:recon_bonn}
    \resizebox{\columnwidth}{!}{
    \setlength{\tabcolsep}{0.6mm}
    \begin{tabular}{lllcccccc}
        \toprule
        Method & Dynamic & Metric & ball & ball2 & ps\_trk & ps\_trk2 & mv\_box2 & Avg. \\
        \midrule
        \rowcolor{gray!20} \multicolumn{9}{l}{\textit{RGB-D}} \\

        \multirow{3}{*}{ {NICE-SLAM~\cite{zhu2022niceslam}}}
        & \multirow{3}{*}{Static}
        & Acc. [cm] $\downarrow$ & X & 24.30 & 43.11 & 74.92 & 17.56 & 39.97 \\
        & & Comp. [cm] $\downarrow$ & X & 16.65 & 117.95 & 172.20 & 19.81 & 81.25 \\
        & & Comp. Ratio [\%] $\uparrow$ & X & 29.68 & 15.89 & 13.96 & 32.18 & 22.93 \\

        \cmidrule{1-9}
        \multirow{3}{*}{ {CO-SLAM~\cite{wang2023coslam} }}
        & \multirow{3}{*}{Static}
        & Acc. [cm] $\downarrow$ & 10.61 & 14.49 & 26.46 & 26.00 & 12.73 & 18.06 \\
        & & Comp. [cm] $\downarrow$ & 10.65 & 40.23 & 124.86 & 118.35 & 10.22 & 60.86 \\
        & & Comp. Ratio [\%] $\uparrow$ & 34.10 & 3.21 & 2.05 & 2.90 & 39.10 & 16.27 \\

        \cmidrule{1-9}
        \multirow{3}{*}{ {ESLAM~\cite{johari2023eslam} }}
        & \multirow{3}{*}{Static}
        & Acc. [cm] $\downarrow$ & 17.17 & 26.82 & 59.18 & 89.22 & 12.32 & 40.94 \\
        & & Comp. [cm] $\downarrow$ & \cellcolor{lightgreen}9.11 & 13.58 & 145.78 & 186.65 & 10.03 & 73.03 \\
        & & Comp. Ratio [\%] $\uparrow$ & 47.44 & \cellcolor{lightyellow}47.94 & 20.53 & 17.33 & 41.41 & 34.93 \\

        \cmidrule{1-9}
        \multirow{3}{*}{ {Rodyn-SLAM~\cite{jiang2024rodynslam}}}
        & \multirow{3}{*}{Dynamic}
        & Acc.[cm] $\downarrow$ & \cellcolor{lightyellow}10.60 & \cellcolor{lightyellow}13.36 & \cellcolor{lightyellow}10.21 & \cellcolor{lightyellow}13.77 & \cellcolor{lightyellow}11.34 & \cellcolor{lightyellow}11.86 \\
        & & Comp.[cm] $\downarrow$ & \cellcolor{softgreen}7.15 & \cellcolor{softgreen}7.87 & \cellcolor{lightyellow}27.70 & \cellcolor{softgreen}19.87 & \cellcolor{softgreen}6.86 & \cellcolor{lightgreen}13.71 \\
        & & Comp. Rat[\%] $\uparrow$ & \cellcolor{lightyellow}47.58 & 40.91 & \cellcolor{lightyellow}34.13 & \cellcolor{lightgreen}32.59 & \cellcolor{lightyellow}45.37 & \cellcolor{lightyellow}40.12 \\

        \cmidrule{1-9}
        \multirow{3}{*}{{DG-SLAM~\cite{xu2025dgslam}}}
        & \multirow{3}{*}{Dynamic}
        & Acc. [cm] $\downarrow$ & \cellcolor{lightgreen}7.00 & \cellcolor{lightgreen}5.80 & \cellcolor{lightgreen}9.14 & \cellcolor{lightgreen}11.78 & \cellcolor{softgreen}6.56 & \cellcolor{lightgreen}8.06 \\
        & & Comp. [cm] $\downarrow$ & 9.80 & \cellcolor{lightgreen}8.05 & \cellcolor{lightgreen}17.99 & \cellcolor{lightgreen}20.10 & \cellcolor{lightgreen}7.61 & \cellcolor{lightyellow}15.46 \\
        & & Comp. Ratio [\%] $\uparrow$ & \cellcolor{lightgreen}49.46 & \cellcolor{softgreen}52.41 & \cellcolor{lightgreen}34.62 & \cellcolor{softgreen}32.81 & \cellcolor{lightgreen}49.02 & \cellcolor{lightgreen}43.67 \\

        \midrule
        \rowcolor{gray!20} \multicolumn{9}{l}{\textit{Monocular}} \\

        \multirow{3}{*}{ {WildGS-SLAM~\cite{zheng2025wildgs-slam}}}
        & \multirow{3}{*}{Dynamic}
        & Acc. [cm] $\downarrow$ & 29.82 & 22.57 & 18.98 & 25.76 & 13.48 & 22.12 \\
        & & Comp. [cm] $\downarrow$ & 36.04 & 23.02 & 30.13 & 40.15 & 15.96 & 29.06 \\
        & & Comp. Ratio [\%] $\uparrow$ & 13.40 & 30.62 & 28.75 & 20.81  & 35.59 & 25.83 \\

        \cmidrule{1-9}
        \multirow{3}{*}{ \textbf{\oursname{} (Ours)}}
        & \multirow{3}{*}{Dynamic}
        & Acc. [cm] $\downarrow$ & \cellcolor{softgreen}\textbf{5.11} & \cellcolor{softgreen}\textbf{5.78} & \cellcolor{softgreen}\textbf{9.07} & \cellcolor{softgreen}\textbf{10.14} & \cellcolor{lightgreen}\textbf{7.64} & \cellcolor{softgreen}\textbf{7.54} \\
        & & Comp. [cm] $\downarrow$ & \cellcolor{lightyellow}\textbf{9.39} & \cellcolor{lightyellow}\textbf{9.78} & \cellcolor{softgreen}\textbf{17.64} & \cellcolor{lightyellow}\textbf{20.24} & \cellcolor{lightyellow}\textbf{7.82} & \cellcolor{softgreen}\textbf{12.87} \\
        & & Comp. Ratio [\%] $\uparrow$ & \cellcolor{softgreen}\textbf{50.10} & \cellcolor{lightgreen}\textbf{51.21} & \cellcolor{softgreen}\textbf{35.42} & \cellcolor{lightyellow}\textbf{32.18} & \cellcolor{softgreen}\textbf{49.50} & \cellcolor{softgreen}\textbf{43.68} \\

        \bottomrule
    \end{tabular}
    }
\end{table}

\subsection{Evaluation of tracking performance}


\begin{table}[h]
    \centering
    \caption{Camera tracking results on several dynamic scene sequences in the \textbf{TUM} dataset. 
    ``X'' and ``-'' denote the tracking failures and absence of mention. The metric is Absolute Trajectory Error (ATE) and the unit is [cm].}
    \label{tab:track_tum}
    \resizebox{\columnwidth}{!}{
    \begin{tabular}{lccccccc}
        \toprule
        \textbf{Method} & \texttt{f3/w\_r} & \texttt{f3/w\_x} & \texttt{f3/w\_s} & \texttt{f3/s\_x}  & \textbf{Avg.} \\
        \midrule
        \rowcolor{gray!20} \multicolumn{8}{l}{\textit{RGB-D}} \\
        ORB-SLAM3 \cite{campos2021orbslam3} 
        & 68.7  & 28.1  & 2.0  & \cellcolor{lightyellow}1.0   & 17.1 \\
        ReFusion \cite{palazzolo2019refusion} 
        & -  & 9.9  & 1.7  & 4.0  & 5.2 \\
        iMAP* \cite{sucar2021imap} 
        & 139.5  & 111.5  & 137.3  & 23.6   & 89.5 \\
        NICE-SLAM* \cite{zhu2022niceslam} 
        & X  & 113.8  & 88.2  & 7.9   & 54.2 \\
        Vox-Fusion \cite{yang2022voxfusion} 
        & X  & 146.6  & 109.9  & 3.8    & 71.6 \\
        Co-SLAM \cite{wang2023coslam} 
        & 52.1  & 51.8  & 49.5  & 7.6  & 32.6 \\
        ESLAM \cite{johari2023eslam} 
        & 90.4  & 45.7  & 93.6  & 7.6   & 48.0 \\
        Rodyn-SLAM \cite{jiang2024rodynslam} 
        & 7.8  & 8.3  & 1.7  & 5.1  & 5.3 \\
        SplaTAM \cite{keetha2024splatam} 
        & 100.4  & 218.3  & 115.2  & 1.7 & 74.4 \\
        DG-SLAM \cite{xu2025dgslam} 
        & 4.3  & 1.6  & \cellcolor{lightyellow}0.6  & \cellcolor{lightyellow}1.0  & 2.2 \\
        DGS-SLAM \cite{kong2024dgs-slam} 
        & - & 4.1 & \cellcolor{lightyellow}0.6 & -& 2.4 \\
        DDN-SLAM (RGB-D) \cite{li2024ddn-slam} 
        & \cellcolor{lightyellow}3.9 & \cellcolor{lightyellow}1.4 & 1.0 & - & 2.1 \\
        GS-SLAM (RGB-D) \cite{matsuki2024gs-slam} 
        & 33.5  & 37.7  & 8.4  & 2.7   & 15.5 \\
        DROID-VO \cite{teed2021droidslam} 
        & 10.0  & 1.7  & 0.7  & 1.1    & 3.3 \\
        \midrule
        \rowcolor{gray!20} \multicolumn{8}{l}{\textit{Monocular}} \\
        DROID-VO \cite{teed2021droidslam} 
        & 6.5  & 1.9  & 1.7  & 1.6    & 2.7 \\
        GS-SLAM (RGB) \cite{matsuki2024gs-slam} 
        & 40.8  & 18.3  & 1.2  & 5.8    & 32.97 \\ 
        Splat-SLAM \cite{sandstrom2024splat-slam} 
        & \cellcolor{lightyellow}3.9 & \cellcolor{lightgreen}1.3  & 2.3  & -   & 2.5 \\
        DDN-SLAM (RGB) \cite{li2024ddn-slam} 
        & 8.9 & 2.8 & 2.5 & -  & 4.73 \\
        GO-SLAM \cite{zhang2023goslam} 
        & 4.1  & 1.6  & 0.8  & 1.1    & \cellcolor{lightyellow}2.05 \\
        WildGS-SLAM \cite{zhang2023goslam} 
        & \cellcolor{lightgreen}3.3  &  \cellcolor{lightgreen}1.3  &  \cellcolor{softgreen}0.4  &  \cellcolor{lightgreen}0.9 &  \cellcolor{lightgreen}1.5  \\
        CUT3R+BA \cite{wang2025cut3r} 
        & 6.8  & 2.6  & 2.2  & 1.9  & 3.4 \\
        \midrule
        \textbf{\oursname{} (Ours)} 
        & \cellcolor{softgreen}\textbf{2.7} & \cellcolor{softgreen}\textbf{1.2} & \cellcolor{lightgreen}\textbf{0.5} & \cellcolor{softgreen}\textbf{0.8} & \cellcolor{softgreen}\textbf{1.3} \\
        \bottomrule
    \end{tabular}
    }
\end{table}
\makeatletter
\def\heavyrulewidth{1.5pt} 
\makeatother


\begin{table}[t]
  \centering
  \caption{Camera tracking performance on the Wild-SLAM MoCap Dataset~\cite{zheng2025wildgs-slam} (ATE $\downarrow$ [cm]).}
  \label{tab:track_wild}
  \normalsize
  \resizebox{0.475\textwidth}{!}{%
  \begin{tabular}{lcccccc}
  \toprule
  \textbf{Method} & \texttt{ANYmal2} & \texttt{Racket} & \texttt{Table1} & \texttt{Table2} & \texttt{Umbre.} & \textbf{Avg.} \\
  \hline
  \rowcolor{gray!20}
  \multicolumn{7}{l}{\textit{\mbox{RGB-D}}} \\
    ReFusion~\cite{palazzolo2019refusion} 
    & 5.6 & 10.4 & 99.1 & 101.0 & 10.7 & 45.36 \\
    DynaSLAM~\cite{bescos2018dynaslam} 
    & \cellcolor{lightyellow}0.5 & \cellcolor{lightyellow}0.8 & 1.2 & 34.8 & 34.7 & 14.40 \\
    \hline
    \rowcolor{gray!20}
    \multicolumn{7}{l}{\textit{Monocular}} \\
    DROID-SLAM~\cite{teed2021droidslam} 
    & 4.7 & 1.5 & 48.0 & 95.6 & 3.8 & 30.72 \\
    DynaSLAM~\cite{bescos2018dynaslam} 
    & \cellcolor{lightyellow}0.5 &\cellcolor{lightgreen} 0.6 & 1.8 & 42.1 & 1.2 & 9.24 \\
    MonST3R~\cite{zhang2024monst3r} 
    & 21.6 & 13.2 & 4.8 & 33.7 & 5.5 & 15.76 \\
    MegaSaM~\cite{li2025megasam} 
    & 2.7 & 1.6 &\cellcolor{lightyellow} 1.0 & 9.4 &\cellcolor{lightyellow} 0.6 &\cellcolor{lightyellow} 3.06 \\
    WildGS-SLAM 
    & \cellcolor{softgreen}0.3 & \cellcolor{softgreen}0.4 & 0.6 & \cellcolor{lightyellow}1.3 & \cellcolor{softgreen}0.2 & \cellcolor{lightgreen}0.56 \\
    $\pi^3_{mos}$-SLAM~\cite{zhong2025pi3mos-slam} 
    & \cellcolor{lightgreen}0.4 & \cellcolor{softgreen}0.4 & \cellcolor{softgreen}0.2 & \cellcolor{lightgreen}1.1 & \cellcolor{softgreen}0.2 & \cellcolor{softgreen}0.46 \\
    CUT3R+BA~\cite{wang2025cut3r} 
    & 8.9 & 5.2 & 4.8 & 7.1 & 4.4 & 6.08 \\
    \midrule
    \textbf{\oursname{} (Ours)} 
    & \cellcolor{softgreen}\textbf{0.3} & \cellcolor{softgreen}\textbf{0.4} & \cellcolor{lightgreen}\textbf{0.4} & \cellcolor{softgreen}\textbf{0.9} & \cellcolor{lightgreen}\textbf{0.3} & \cellcolor{softgreen}\textbf{0.46} \\
  \bottomrule
  \end{tabular}%
  }
\end{table}

To effectively showcase the tracking performance of our \oursname{} method, we conduct comparative analyses within highly dynamic, slightly dynamic and static environments. The comparison methods encompass classical SLAM systems such as ORB-SLAM3 \cite{campos2021orbslam3}, ReFusion \cite{palazzolo2019refusion}, Co-fusion \cite{runz2017cofusion}, MID-fusion \cite{xu2019midfusion}, EM-fusion \cite{strecke2019emfusion}, and widely recognized NeRF-based SLAM methods such as iMAP \cite{sucar2021imap}, NICE-SLAM \cite{zhu2022niceslam}, Vox-Fusion \cite{yang2022voxfusion}, Co-SLAM \cite{wang2023coslam}, ESLAM \cite{johari2023eslam}. We also incorporate a comparison with the newly proposed dynamic RGB-D SLAM systems, including DGS-SLAM \cite{kong2024dgs-slam}, Rodyn-SLAM \cite{jiang2024rodynslam} and DG-SLAM \cite{xu2025dgslam}. Additionally, monocular SLAM approaches such as GO-SLAM \cite{zhang2023goslam}, NICER-SLAM \cite{zhu2024nicerslam} and GS-SLAM \cite{matsuki2024gs-slam} are also compared with. DDN-SLAM \cite{li2024ddn-slam} proposes specific designs for dynamic scenes.  

We report the results on four sequences from TUM dataset as summarized in Table~\ref{tab:track_tum}. Our proposed system achieves consistently strong tracking performance, attributed to the integration of semantic-aware multi-view consistency and the powerful coarse-to-fine camera tracking algorithm. To further assess the generalizability and robustness of our method, we conduct extensive experiments on 
Wild-Mocap dataset~\cite{zheng2025wildgs-slam} in Table~\ref{tab:track_wild}. Across these challenging scenarios, our method consistently surpasses state-of-the-art methods, highlighting its effectiveness and reliability in real-world navigation applications. 
The qualitative comparison of rendered images and depths is illustrated in Figure~\ref{fig:supp_tum} and Figure~\ref{fig:supp_bonn}, where our method demonstrates superior performance in distractor removal on both TUM and Bonn datasets. 

\subsection{Ablation study}
         



\begin{table}[th]
\centering
\caption{\textbf{\oursname{}} Ablation studies of different components on Bonn and TUM. We report ATE [cm].}
\label{tab:ablation}
\resizebox{0.9\columnwidth}{!}{
\setlength{\tabcolsep}{1.0mm}
    \begin{tabular}{lcc}
        \midrule
        & \textbf{Bonn} & \textbf{TUM} \\
        \midrule
        (a) w/o long-term dynamic mask & 
        3.61 & \cellcolor{lightyellow}1.98 \\
        (b) w/o semantic-aware multi-view consistency & 
        \cellcolor{lightyellow}3.57 & \cellcolor{lightgreen}1.65 \\
        (c) w/o semantic-guided occlusion-aware keyframes & 
        \cellcolor{lightgreen}3.31 & 2.1 \\
        \textbf{\oursname{} (Ours)} & 
        \cellcolor{softgreen}\textbf{2.9} & \cellcolor{softgreen}\textbf{1.3} \\
        \midrule
    \end{tabular}
}
\end{table}

To evaluate the efficacy of the proposed algorithm in our system, we conducted ablation studies across two datasets TUM and Bonn where both sequences consist of predictable motions such as pedestrians and unpredictable object motions such as chairs and balloons. 
We calculated the average ATE for tacking performance to illustrate the impact of various components on the overall system capability. 
As the results shown in Table~\ref{tab:ablation}, the findings affirm the effectiveness of all proposed methods in improving camera tracking. 
The strategies of semantic-aware multi-view consistency and semantic-guided keyframe selection substantially enhance the quality of mapping, which in turn exerts a notable positive impact on tracking accuracy.

\subsection{Time consumption analysis}
As presented in Table~\ref{tab:runtime}, we report the per-frame computational cost for both tracking and mapping, excluding the time required for semantic segmentation. All results were obtained under an identical experimental protocol, performing 40 iterations for tracking and 60 iterations for mapping. Leveraging the high efficiency of \cite{teed2021droidslam} for pose estimation and the fast rendering capabilities of 3D Gaussian Splatting, our method achieves superior tracking speed compared to existing approaches. 

\begin{table}[t]
\centering
\caption{Run-time comparison on \texttt{TUM f3/w\_x}.}
\label{tab:runtime}
\resizebox{0.8\columnwidth}{!}{%
\setlength{\tabcolsep}{1.0mm}
    \begin{tabular}{lcc}
        \midrule
         & \textbf{Tracking} [ms]$\downarrow$ & \textbf{Mapping} [ms]$\downarrow$ \\
        \midrule
        NICE-SLAM~\cite{zhu2022niceslam} 
        & 3186.2 & 1705.1 \\
        ESLAM~\cite{johari2023eslam} 
        & \cellcolor{lightyellow}2045.9 & 1641.4 \\
        Point-SLAM~\cite{sandstrom2023pointslam} 
        & 2279.5 & \cellcolor{lightyellow}1544.4 \\
        DG-SLAM~\cite{xu2025dgslam}
        & \cellcolor{softgreen}89.2 & \cellcolor{lightgreen}549.3 \\
        \midrule
        \textbf{\oursname (Ours)} 
        & \cellcolor{lightgreen}\textbf{97.5} & \cellcolor{softgreen}\textbf{216.7} \\
        \midrule
    \end{tabular}
}
\end{table}

\section{Conclusion}

In this paper, we propose \oursname{}, a robust semantic-aware Gaussian Splatting SLAM from monocular video under dynamic environments. Via robust long-term dynamic mask strategy for coarse-to-fine tracking algorithm, our system significantly advances the accuracy and robustness of pose estimation within dynamic scenes. The proposed semantic-aware multi-view consistency and semantic-guided keframes for occlusion inpainting stratey effectively improves the quality of reconstructed maps and rendering images. We demonstrate its effectiveness in achieving superior results in camera pose estimation and scene reconstruction in dynamic scenes. 

\section*{Acknowledgments}
This work is supported by the Centre for Perceptual and Interactive Intelligence (CPII) Ltd., a CUHK-led under the InnoHK scheme of Innovation and Technology Commission of the Hong Kong Special Administrative Region Government. This work is also partially supported by the HK RGC AoE under AoE/E-407/24-N.

\bibliographystyle{IEEEtran}
\bibliography{slam-bib}



\vfill

\end{document}